\def\year{2017}\relax
\documentclass[letterpaper]{article}
\usepackage{aaai17}
\usepackage{times}
\usepackage{helvet}
\usepackage{courier}

\usepackage{amsmath}
\usepackage[utf8]{inputenc}
\usepackage{url}
\usepackage{graphicx}
\usepackage{amsmath}

\begin{document}

\title{ConceptNet 5.5: An Open Multilingual Graph of General Knowledge}
\author{Robyn Speer\\
    Luminoso Technologies, Inc.\\
    675 Massachusetts Avenue\\
    Cambridge, MA 02139
\And
    Joshua Chin\\
    Union College\\
    807 Union St.\\
    Schenectady, NY 12308\\
\And
    Catherine Havasi\\
    Luminoso Technologies, Inc.\\
    675 Massachusetts Avenue\\
    Cambridge, MA 02139
}

\maketitle

\section{Abstract}\label{abstract}

Machine learning about language can be improved by supplying it with specific knowledge and sources of external information. We present here a new version of the linked open data resource ConceptNet that is particularly well suited to be used with modern NLP techniques such as word embeddings.

ConceptNet is a knowledge graph that connects words and phrases of natural language with labeled edges. Its knowledge is collected from many sources that include expert-created resources, crowd-sourcing, and games with a purpose. It is designed to represent the general knowledge involved in understanding language, improving natural language applications by allowing the application to better understand the meanings behind the words people use.

When ConceptNet is combined with word embeddings acquired from distributional semantics (such as word2vec), it provides applications with understanding that they would not acquire from distributional semantics alone, nor from narrower resources such as WordNet or DBPedia. We demonstrate this with state-of-the-art results on intrinsic evaluations of word relatedness that translate into improvements on applications of word vectors, including solving SAT-style analogies.

\section{Introduction}\label{introduction}

ConceptNet is a knowledge graph that connects words and phrases of
natural language (\emph{terms}) with labeled, weighted edges
(\emph{assertions}). The original release of ConceptNet \cite{liu2004conceptnet}
was intended as a parsed representation of Open Mind Common Sense
\cite{singh2002omcs}, a crowd-sourced knowledge project. This paper
describes the release of ConceptNet 5.5, which has expanded to include
lexical and world knowledge from many different sources in many
languages.

ConceptNet represents relations between words such as:

\begin{itemize}
    \item A \emph{net} is used for \emph{catching fish}.
    \item ``\emph{Leaves}'' is a form of the word ``\emph{leaf}''.
    \item The word \emph{cold} in English is \emph{studený} in Czech.
    \item O \emph{alimento} é usado para \emph{comer} [Food is used for eating].
\end{itemize}

In this paper, we will concisely represent assertions such as the above as triples of
their start node, relation label, and end node: the assertion that ``a dog has
a tail'' can be represented as (\emph{dog}, \emph{HasA}, \emph{tail}).

ConceptNet also represents links between knowledge resources. In addition to
its own knowledge about the English term \emph{astronomy}, for example,
ConceptNet contains links to URLs that define \emph{astronomy} in WordNet,
Wiktionary, OpenCyc, and DBPedia.

The graph-structured knowledge in ConceptNet can be particularly useful to NLP
learning algorithms, particularly those based on word embeddings, such as
\cite{mikolov2013word2vec}. We can use ConceptNet to build semantic spaces that
are more effective than distributional semantics alone.

The most effective semantic space is one that learns from both distributional
semantics and ConceptNet, using a generalization of the ``retrofitting'' method
\cite{faruqui2015retrofitting}. We call this hybrid semantic space ``ConceptNet
Numberbatch'', to clarify that it is a separate artifact from ConceptNet
itself.

ConceptNet Numberbatch performs significantly better than other systems across
many evaluations of word relatedness, and this increase in performance
translates to improvements on downstream tasks such as analogies.  On a
corpus of SAT-style analogy questions \cite{turney2006lra}, its accuracy of
56.1\% outperforms other systems based on word embeddings and ties the
previous best overall system, Turney's LRA. This level of accuracy is only
slightly lower than the performance of the average human test-taker.

Building word embeddings is not the only application of ConceptNet, but it is a
way to apply ConceptNet that achieves clear benefits and is compatible with
ongoing research in distributional semantics.

After introducing related work, we will begin by describing ConceptNet 5.5 and
its features, show how to use ConceptNet alone as a semantic space and a
measure of word relatedness, and then proceed to describe and evaluate the
hybrid system ConceptNet Numberbatch on these various semantic tasks.

\section{Related Work}

ConceptNet is the knowledge graph version of the Open Mind
Common Sense project \cite{singh2002omcs}, a common sense knowledge base
of the most basic things a person knows. It was last published as version
5.2 \cite{speer2013conceptnet}.

Many projects strive to create lexical resources of general knowledge.  Cyc
\cite{lenat1989cyc} has built an ontology of common-sense knowledge in
predicate logic form over the decades. DBPedia \cite{auer2007dbpedia} extracts
knowledge from Wikipedia infoboxes, providing a large number of facts, largely
focused on named entities that have Wikipedia articles. The Google Knowledge
Graph \cite{singhal2012googleblog} is perhaps the largest and most general
knowledge graph, though its content is not freely available. It focuses largely
on named entities that can be disambiguated, with a motto of ``things, not
strings''.

ConceptNet's role compared to these other resources is to provide a
sufficiently large, free knowledge graph that focuses on the common-sense
meanings of words (not named entities) as they are used in natural language.
This focus on words makes it particularly compatible with the idea of
representing word meanings as vectors.

Word embeddings represent words as dense unit vectors of real numbers, where
vectors that are close together are semantically related. This representation
is appealing because it represents meaning as a continuous space, where
similarity and relatedness can be treated as a metric. Word embeddings are
often produced as a side-effect of a machine learning task, such as predicting
a word in a sentence from its neighbors.  This approach to machine learning
about semantics is sometimes referred to as \emph{distributional semantics} or
\emph{distributed word representations}, and it contrasts with the
knowledge-driven approach of semantic networks or knowledge graphs.

Two prominent matrices of embeddings are the word2vec embeddings trained on 100
billion words of Google News using skip-grams with negative sampling
\cite{mikolov2013word2vec}, and the GloVe 1.2 embeddings trained on 840 billion
words of the Common Crawl \cite{pennington2014glove}. These matrices are
downloadable, and we will be using them both as a point of comparison and as
inputs to an ensemble. \citeauthor{levy2015embeddings}
\shortcite{levy2015embeddings} evaluated multiple embedding techniques and the
effects of various explicit and implicit hyperparameters, produced their own
performant word embeddings using a truncated SVD of words and their contexts,
and provided recommendations for the engineering of word embeddings.

Holographic embeddings \cite{nickel2015holographic} are embeddings learned from
a labeled knowledge graph, under the constraint that a circular correlation of
these embeddings gives a vector representing a relation. This representation
seems extremely relevant to ConceptNet. In our attempt to implement it on
ConceptNet so far, it has converged too slowly to experiment with, but this
could be overcome eventually with some optimization and additional computing
power.

\section{Structure of ConceptNet}\label{structure-of-conceptnet}

\begin{figure}[t]
\centering
\includegraphics[width=3.3in]{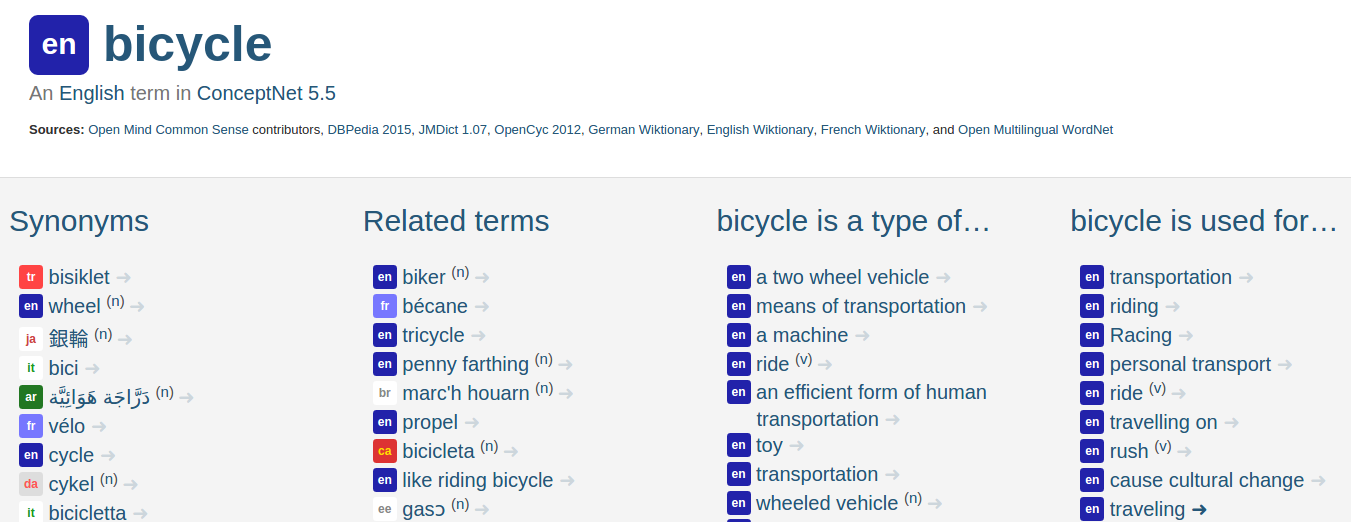}
\caption{
    ConceptNet's browsable interface (conceptnet.io) shows
    facts about the English word ``bicycle''.
}
\label{screenshot}
\end{figure}

\subsection{Knowledge Sources}\label{knowledge-sources}

ConceptNet 5.5 is built from the following sources:

\begin{itemize}
\item
  Facts acquired from Open Mind Common Sense (OMCS) \cite{singh2002omcs}
  and sister projects in other languages \cite{anacleto2006portuguese}
\item
  Information extracted from parsing Wiktionary, in multiple languages,
  with a custom parser (``Wikiparsec'')
\item
  ``Games with a purpose'' designed to collect common knowledge
  \cite{vonahn2006verbosity} \cite{nakahara2011nadya} \cite{kuo2009petgame}
\item
  Open Multilingual WordNet \cite{bond2013linking}, a linked-data
  representation of WordNet \cite{miller1998wordnet} and its parallel
  projects in multiple languages
\item
  JMDict \cite{breen2004jmdict}, a Japanese-multilingual dictionary
\item
  OpenCyc, a hierarchy of hypernyms provided by
  Cyc \cite{lenat1989cyc}, a system that represents common sense knowledge in predicate logic
\item
  A subset of DBPedia \cite{auer2007dbpedia}, a network of facts
  extracted from Wikipedia infoboxes
\end{itemize}

With the combination of these sources, ConceptNet contains over 21
million edges and over 8 million nodes. Its English vocabulary contains
approximately 1,500,000 nodes, and there are 83 languages in which it
contains at least 10,000 nodes.

The largest source of input for ConceptNet is Wiktionary, which provides
18.1 million edges and is mostly responsible for its large multilingual
vocabulary. However, much of the character of ConceptNet comes from OMCS
and the various games with a purpose, which express many different kinds
of relations between terms, such as \emph{PartOf} (``a wheel is part of
a car'') and \emph{UsedFor} (``a car is used for driving'').

\subsection{Relations}\label{relations}

ConceptNet uses a closed class of selected relations such as \emph{IsA},
\emph{UsedFor}, and \emph{CapableOf}, intended to
represent a relationship independently of the language or the source of
the terms it connects.

ConceptNet 5.5 aims to align its knowledge resources on its core set of 36
relations. These generalized relations are similar in purpose to WordNet's
relations such as \emph{hyponym} and \emph{meronym}, as well as to the qualia
of the Generative Lexicon theory \cite{pustejovsky1991generative}.
ConceptNet's edges are directed, but as a new feature in ConceptNet 5.5,
some relations are designated as being symmetric, such as \emph{SimilarTo}.
The directionality of these edges is unimportant.

The core relations are:
\begin{itemize}
    \item {\bf Symmetric relations}: \emph{Antonym}, \emph{DistinctFrom},
        \emph{EtymologicallyRelatedTo}, \emph{LocatedNear},
        \emph{RelatedTo}, \emph{SimilarTo}, and \emph{Synonym}
    \item {\bf Asymmetric relations}: \emph{AtLocation}, \emph{CapableOf},
        \emph{Causes}, \emph{CausesDesire}, \emph{CreatedBy},
        \emph{DefinedAs}, \emph{DerivedFrom}, \emph{Desires},
        \emph{Entails}, \emph{ExternalURL},
        \emph{FormOf}, \emph{HasA}, \emph{HasContext},
        \emph{HasFirstSubevent}, \emph{HasLastSubevent},
        \emph{HasPrerequisite}, \emph{HasProperty}, \emph{InstanceOf},
        \emph{IsA}, \emph{MadeOf}, \emph{MannerOf}, \emph{MotivatedByGoal},
        \emph{ObstructedBy}, \emph{PartOf}, \emph{ReceivesAction},
        \emph{SenseOf}, \emph{SymbolOf}, and \emph{UsedFor}
\end{itemize}

Definitions and examples of these relations appear in a page of the ConceptNet 5.5
documentation\footnote{\url{https://github.com/commonsense/conceptnet5/wiki/Relations}}.

Relations with specific semantics, such as \emph{UsedFor} and
\emph{HasPrerequisite}, tend to connect common words and phrases, while
rarer words are connected by more general relations such as
\emph{Synonym} and \emph{RelatedTo}.

An example of edges in ConceptNet, in a browsable interface that groups them by
their relation expressed in natural English, appears in
Figure~\ref{screenshot}.

\subsection{Term Representation}\label{term-representation}

ConceptNet represents terms in a standardized form. The text is
Unicode-normalized in NFKC form\footnote{\url{http://unicode.org/reports/tr15/}}
using Python's {\tt unicodedata} implementation, lowercased, and split into non-punctuation tokens
using the tokenizer in the Python package \texttt{wordfreq}
\cite{speer2016wordfreq}, which builds on the standard Unicode word
segmentation algorithm. The tokens are joined with underscores, and this text
is prepended with the URI \texttt{/c/lang}, where \emph{lang} is the BCP 47
language code\footnote{\url{https://tools.ietf.org/html/bcp47}} for the language the term is in. As an example, the English term
``United States'' becomes \texttt{/c/en/united\_states}.

Relations have a separate namespace of URIs prefixed with \texttt{/r}, such as
\texttt{/r/PartOf}. These relations are given artificial names in English, but
apply to all languages. The statement that was obtained in Portuguese as ``O
\emph{alimento} é usado para \emph{comer}'' is still represented with the
relation \texttt{/r/UsedFor}.

The most significant change from ConceptNet 5.4 and earlier is in the
representation of terms. ConceptNet 5.4 required terms in English to be in
lemmatized form, so that, for example, ``United States'' had to be represented
as \texttt{/c/en/unite\_state}. In this representation, ``drive'' and
``driving'' were the same term, allowing the assertions (\emph{car},
\emph{UsedFor}, \emph{driving}) and (\emph{drive}, \emph{HasPrerequisite},
\emph{have license}) to be connected. ConceptNet 5.5 removes the lemmatizer,
and instead relates inflections of words using the \emph{FormOf} relation. The
two assertions above are now linked by the third assertion (\emph{driving},
\emph{FormOf}, \emph{drive}), and both ``driving'' and ``drive'' can be looked
up in ConceptNet.

\subsection{Vocabulary}\label{vocabulary}

When building a knowledge graph, the decision of what a node should
represent has significant effects on how the graph is used. It also has
implications that can make linking and importing other resources
non-trivial, because different resources make different decisions about
their representation.

In ConceptNet, a node is a word or phrase of a natural language, often a common
word in its undisambiguated form. The word ``lead'' in English is a term in
ConceptNet, represented by the URI \texttt{/c/en/lead}, even though it has
multiple meanings. The advantage of ambiguous terms is that they can be
extracted easily from natural language, which is also ambiguous. This ambiguous
representation is equivalent to that used by systems that learn distributional
semantics from text.

ConceptNet's representation allows for more specific, disambiguated
versions of a term. The URI \texttt{/c/en/lead/n} refers to noun senses
of the word ``lead'', and is effectively included within
\texttt{/c/en/lead} when searching or traversing ConceptNet, and
linked to it with the implicit relation \emph{SenseOf}. Many data
sources provide information about parts of speech, allowing us to use
this as a common representation that provides a small amount of
disambiguation. Further disambiguation is allowed by the URI structure,
but not currently used.

\subsection{Linked Data}

ConceptNet imports knowledge from some other systems, such as WordNet, into its
own representation. These other systems have their own target vocabularies that
need to be aligned with ConceptNet, which is usually an underspecified,
many-to-many alignment.

A term that is imported from another knowledge graph will be connected to
ConceptNet nodes via the relation \emph{ExternalURL}, pointing to an absolute
URL that represents that term in that external resource. This newly-introduced
relation preserves the provenance of the data and enables looking up what the
untransformed data was. ConceptNet terms can also be represented as absolute
URLs, so this allows ConceptNet to connect bidirectionally to the broader
ecosystem of Linked Open Data.

\section{Applying ConceptNet to Word Embeddings}

\subsection{Computing ConceptNet Embeddings Using PPMI}

We can represent the ConceptNet graph as a sparse, symmetric term-term matrix.
Each cell contains the sum of the weights of all edges that connect the two
corresponding terms. For performance reasons, when building this matrix, we
prune the ConceptNet graph by discarding terms connected to fewer than three
edges.

We consider this matrix to represent terms and their contexts. In a corpus of
text, the context of a term would be the terms that appear nearby in the text;
here, the context is the other nodes it is connected to in ConceptNet. We can
calculate word embeddings directly from this sparse matrix by following the
practical recommendations of \citeauthor{levy2015embeddings}
\shortcite{levy2015embeddings}.

As in Levy et al., we determine the pointwise mutual information of the matrix
entries with context distributional smoothing, clip the negative values to
yield positive pointwise mutual information (PPMI), reduce the dimensionality
of the result to 300 dimensions with truncated SVD, and combine the terms and contexts
symmetrically into a single matrix of word embeddings.

This gives a matrix of word embeddings we call ConceptNet-PPMI.  These
embeddings implicitly represent the overall graph structure of ConceptNet, and
allow us to compute the approximate connectedness of any pair of nodes.

We can expand ConceptNet-PPMI to restore the nodes that we pruned away,
assigning them vectors that are the average of their neighboring nodes.

\subsection{Combining ConceptNet with Distributional Word Embeddings}

Having created embeddings from ConceptNet alone, we would now like to create
a more robust set of embeddings that represents both ConceptNet and distributional
word embeddings learned from text.

Retrofitting \cite{faruqui2015retrofitting} is a process that adjusts an
existing matrix of word embeddings using a knowledge graph. Retrofitting
infers new vectors $q_i$ with the objective of being close to their original
values, $\hat{q_i}$, and also close to their neighbors in the graph with edges $E$,
by minimizing this objective function:
$$\Psi(Q) = \sum_{i=1}^{n}\left[
    \alpha_i \lVert q_i - \hat{q_i} \rVert ^2 + \sum_{(i, j) \in E} \beta_{ij} \lVert q_i - q_j \rVert ^2
\right] $$

\citeauthor{faruqui2015retrofitting} give a simple iterative process to minimize
this function over the vocabulary of the original embeddings.

The process of ``expanded retrofitting'' \cite{speer2016ensemble} can optimize
this objective over a larger vocabulary, including terms from the knowledge
graph that do not appear in the vocabulary of the word embeddings. This
effectively sets $\alpha_i = 0$ for terms whose original values are undefined.
We set $\beta_{ij}$ according to the weights of the edges in ConceptNet.

The particular benefit of expanded retrofitting to ConceptNet is that it can
benefit from the multilingual connections in ConceptNet. It learns more about
English words via their translations in other languages, and also gives these
foreign-language terms useful embeddings in the same space as the English
terms. The effect is similar to the work of \citeauthor{xiao2014distributed}
\shortcite{xiao2014distributed}, who also propagate multilingual embeddings
using crowd-sourced Wiktionary entries.

We add one more step to retrofitting, which is to subtract the mean of the
vectors that result from retrofitting, then re-normalize them to unit vectors.
Retrofitting has a tendency to move all vectors closer to the vectors for
highly-connected terms such as ``person''. Subtracting the mean helps to ensure
that terms remain distinguishable from each other.

\subsection{Combining Multiple Sources of Embeddings}

Retrofitting can be applied to any existing matrix of word embeddings, without
needing access to the data that was used to train them. This is particularly
useful because it allows building on publicly-released matrices of embeddings
whose input data is unavailable or difficult to acquire.

As described in the ``Related Work'' section, word2vec and GloVe both provide
recommended pre-trained matrices. These matrices represent somewhat different
domains of text and have complementary strengths, and the way that we can
benefit from them the most is by taking both of them as input.

To do this, we apply retrofitting to both matrices, then find a globally
linear projection that aligns the results on their common vocabulary.
This process was inspired by \citeauthor{zhao2015learning} \shortcite{zhao2015learning}.
We find the projection by concatenating the columns of the matrices and reducing
them to 300 dimensions using truncated SVD. We then use this alignment to infer
compatible embeddings for terms that are missing from one of the vocabularies.

In ongoing work, we are experimenting with additionally including
distributional word embeddings from corpora of non-English text in this merger.
Preliminary results show that this improves the multilingual performance of the
embeddings.

After retrofitting and merging, we have a labeled matrix of word embeddings
whose vocabulary is derived from word2vec, GloVe, and the pruned ConceptNet
graph. As in ConceptNet-PPMI, we re-introduce all the nodes from ConceptNet by
looking up and averaging their neighboring nodes.

\section{Evaluation}

To compare the performance of fully-built systems of word embeddings, we will
first compare their results on intrinsic evaluations of word relatedness, then
apply the word embeddings to the downstream tasks of solving proportional
analogies and choosing the sensible ending to a story, to evaluate whether
better embeddings translate to better performance on semantic tasks.

The hybrid system described above is the system we name ConceptNet Numberbatch,
with the version number 16.09 indicating that it was built in September 2016.
We now compare results from ConceptNet Numberbatch 16.09 to other systems that
make their word embeddings available, both those that were used in building
ConceptNet Numberbatch and a recently-released system, LexVec, that was not.
The systems we evaluate are:

\begin{itemize}
    \item word2vec SGNS \cite{mikolov2013word2vec}, trained on Google News text
    \item GloVe 1.2 \cite{pennington2014glove}, trained on the Common Crawl
    \item LexVec \cite{salle2016lexvec}, trained on the English Wikipedia and NewsCrawl 2014
    \item ConceptNet-PPMI, described here and trained on ConceptNet 5.5 alone
    \item ConceptNet Numberbatch 16.09, the hybrid of ConceptNet 5.5, word2vec, and GloVe described here
\end{itemize}

\subsection{Evaluations of Word Relatedness}
\label{intrinsic-evaluations}

One way to evaluate the intrinsic performance of a semantic space is to ask it
to rank the relatedness of pairs of words, and compare its judgments to human
judgments.\footnote{It is sometimes important to distinguish \emph{similarity}
from \emph{relatedness}. For example, the term ``coffee'' is related to
``mug'', but coffee is not \emph{similar} to a mug. What a machine can learn
from the connectivity of ConceptNet is focused on relatedness.} If one word in
a pair is out-of-vocabulary, the pair is assumed to have a relatedness of 0. A
good semantic space will provide a ranking of relatedness that is highly
correlated with the human gold-standard ranking, as measured by its Spearman
correlation ($\rho$).

Many gold standards of word relatedness are in common use. Here, we focus on
MEN-3000 \cite{bruni2014men}, a large crowd-sourced ranking of common words; RW
\cite{luong2013rw}, a ranking of rare words; WordSim-353 \cite{finkelstein2001ws},
a smaller evaluation that has been used as a benchmark for many methods; and MTurk-771
\cite{halawi2012mturk}, another crowd-sourced evaluation of a variety of words.

To avoid manually overfitting by designing our semantic space around a
particular evaluation, we experimented using smaller development sets, holding
out some test data until it was time to include results in this paper:

\begin{itemize}
\item
    MEN-3000 is already divided into a 2000-item development set and a
    1000-item test set. We use the results from the test set as the final results.
\item
    RW has no standard dev/test breakdown. We sampled 2/3 of its items as
    a development set and held out the other 1/3 (every third line of the file,
    starting with the third).
\item
    We used all of WordSim-353 in development. We examine its results both in
    English and in its Spanish translation \cite{hassan2009crosslingual}.
\item
    We did not use MTurk-771 in development, holding out the entire set
    as a final test, showing that ConceptNet Numberbatch performs well on a
    previously-unseen evaluation.
\end{itemize}

We use the Spanish WordSim-353 as an example of a prominent non-English evaluation,
indicating that expanded retrofitting is sufficient to learn vectors for
non-English languages, even when all the distributional semantics takes place in
English. However, a thorough multilingual evaluation is beyond the scope of
this paper; the systems we compare to have only made English vectors available,
and it would add considerable complexity to the evaluation to reproduce other
systems of multilingual embeddings, accounting for their various ways of
handling morphology and OOV words.

\subsection{Solving SAT-style Analogies}

Proportional analogies are statements of the form ``$a_1$ is to $b_1$ as $a_2$
is to $b_2$''. The task of filling in missing values of a proportional analogy
was common until recently on standardized tests such as the SAT. Now, it is
popular as a way to show that a semantic space can approximate relationships
between words, even without taking explicit relationships into account.

Much of the groundwork for evaluating systems' ability to solve proportional
analogies was laid by Peter Turney, including his method of Latent Relational
Analysis \cite{turney2006lra}, which was quite effective at solving
proportional analogies by repeatedly searching the Web for the words involved
in them. A newer method called SuperSim \cite{turney2013supersim} does
not require Web searching. These methods are evaluated on a dataset of 374 SAT
questions that Turney and his collaborators have collected.

Many of the best results on this evaluation have been achieved by Turney in his
own work. One interesting system not by Turney is BagPack
\cite{herdagdelen2009bagpack}, which could learn about analogies either from
unstructured text or from ConceptNet 4.

Solving analogies over word embeddings is often described as comparing the
difference $b_2 - a_2$ to $b_1 - a_1$ \cite{mikolov2013word2vec}, but for the
task of filling in the best pair for $a_2$ and $b_2$, it helps to take
advantage of more of the structure of the question to provide more constraint
than this single comparison.

In a sensible analogy, the words on the right side of the analogy will be
related in some way to the words on the left side, so we should aim for some
amount of relatedness between $a_1$ and $a_2$, and between $b_1$ and $b_2$,
regardless of what the other terms are. Also, in many cases, a satisfying
analogy will still make sense when it is transposed to ``$a_1$ is to $a_2$ as
$b_1$ is to $b_2$''. The analogy ``fire : hot :: ice : cold'', for example, can
be transposed to ``fire : ice :: hot : cold''. Recognizing this structure helps
in picking the best answer to difficult analogy questions.

This gives us three components that we can weigh to evaluate whether a pair
($a_2$, $b_2$) completes an analogy: their separate similarity to $a_1$ and
$b_1$, the dot product of differences between the pairs, and the dot product of
differences between the transposed pairs.  The total weight does not matter, so
we can put these together into a vector equation with two free parameters:
\begin{equation*}
    \begin{split}
        s &= a_1 \cdot a_2 + b_1 \cdot b_2\\
          &  + w_1(b_2 - a_2) \cdot (b_1 - a_1)
             + w_2(b_2 - b_1) \cdot (a_2 - a_1)
    \end{split}
\end{equation*}

The appropriate values of $w_1$ and $w_2$ depend on the nature of the
relationships in the analogy questions, and also on how these relationships
appear in the vector space. We optimize these parameters separately for each
system we test, using grid search over a number of possible values so that each
system can achieve its best performance. The grid search is performed on
odd-numbered questions, holding out the even-numbered questions as a test set.

The weights found for ConceptNet Numberbatch 16.09 were $w_1 = 0.2$ and $w_2 =
0.6$.  This indicates, surprisingly, that the comparisons being made by the
transposed form of the analogy were often more important than the directly
stated form of the analogy for choosing the best answer pair.

\subsection{An Evaluation of Common-Sense Stories}
\label{story-evaluation}

The Story Cloze Test \cite{mostafazadeh2016cloze} is a recent evaluation of
semantic understanding that tests whether a method can choose the sensible
ending to a simple story. Prompts consist of four sentences that tell a story,
and two choices are provided for a fifth sentence that concludes the story,
only one of which makes sense.

This task is distinguished by being very challenging for computers but very
easy for humans, because of the extent that it relies on implicit, common sense
knowledge. Most systems that have been evaluated on the Story Cloze Test
score only marginally above the random baseline of 50\%, while human
agreement is near 100\%.

Our preliminary attempt to apply ConceptNet Numberbatch to the Story Cloze Test
is to use a very simple ``bag-of-vectors'' model, by averaging the
embeddings of the words in the sentence and choosing the ending whose average is
closest. This allows us to compare directly to one of the original results presented by
\citeauthor{mostafazadeh2016cloze}, in which a bag of vectors using GenSim's
implementation of word2vec scores 53.9\% on the test set.

This bag-of-vectors model uses no knowledge of how one event might sensibly
follow from another, only which words are related in context. Improving the
score of this model should not be portrayed as actual ``story understanding'',
but it recognizes that sensible stories do not suddenly change topic.

\section{Results and Discussion}
\begin{figure}[t]
\centering
\includegraphics[width=3.3in]{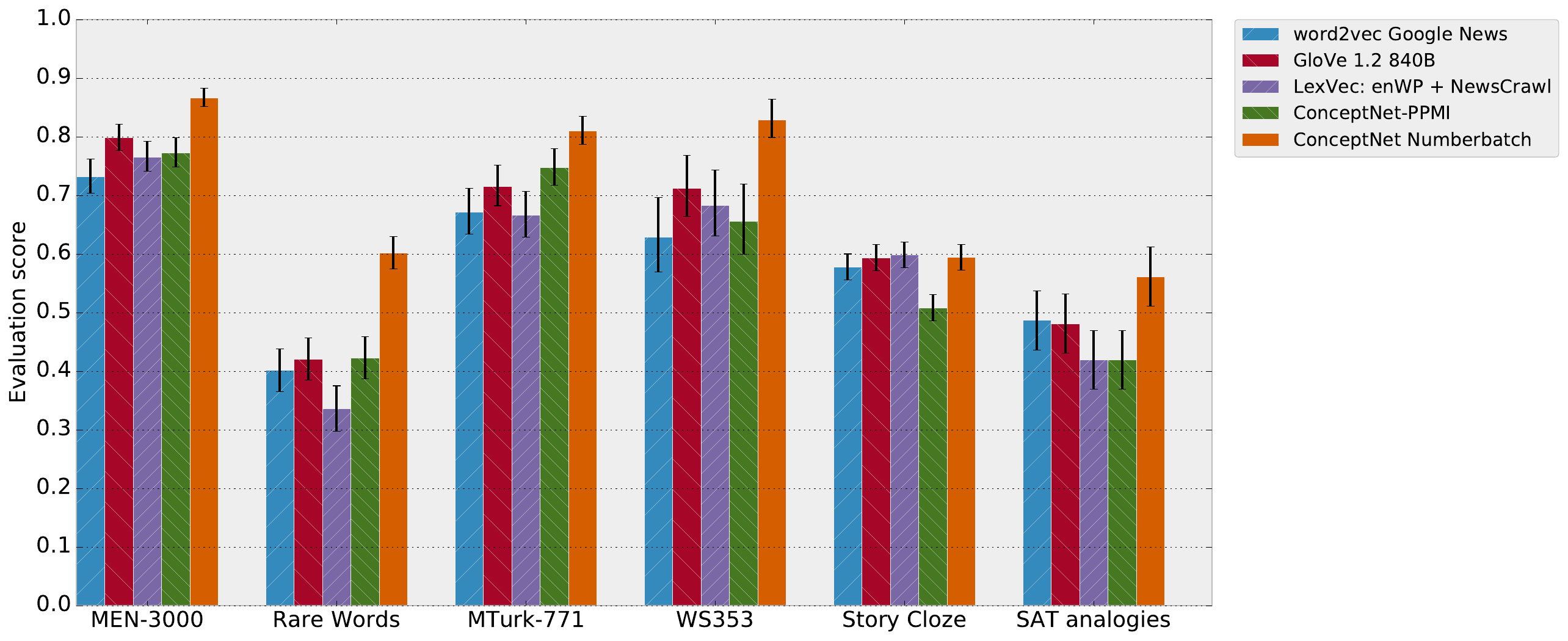}
\caption{
    Performance of word embeddings across multiple evaluations.
    Error bars show 95\% confidence intervals.
}
\label{eval-results}
\end{figure}

\begin{table}[t]
\centering
\begin{tabular}{lrrr}
\bf Evaluation     & \bf Dev & \bf Test & \bf Final \\
\hline
MEN-3000 ($\rho$)              & .859 & .866 & .866 \\
Rare Words ($\rho$)            & .609 & .586 & .601 \\
MTurk-771 ($\rho$)             &  --- & .810 & .810 \\
WordSim-353 ($\rho$)           & .828 &  --- & .828 \\
WordSim-353 Spanish ($\rho$)   & .685 &  --- & .685 \\
Story Cloze Test (acc)         & .604 & .594 & .594 \\
SAT Analogies (acc)            & .535 & .588 & .561
\end{tabular}
\caption{The Spearman correlation ($\rho$) or accuracy (acc) of ConceptNet Numberbatch 16.09, our hybrid system,
on data used in development and data held out for testing.}
\label{eval-table}
\end{table}

\subsection{Word Relatedness}

Figure~\ref{eval-results} compares the performance of the systems we compared
across all evaluations. For word-relatedness evaluations, the Y-axis represents
the Spearman correlation ($\rho$), using the Fisher transformation to compute a
95\% confidence interval that assumes the given word pairs are sampled from an
unobservable larger set \cite{bonett2000sample}. For the analogy and story evaluations,
the Y-axis is simply the proportion of questions answered correctly, with 95\%
confidence intervals calculated using the binomial exact test.

The scores of our system on all these evaluations appear in
Table~\ref{eval-table}, including a development/test breakdown that shows no
apparent overfitting. The ``Final'' column is meant for comparisons to other
papers and used in the graph. It uses the standard test set that other
publications use, if it exists (which is the case for MEN-3000 and Story
Cloze), or all of the data otherwise.

On all of the four word-relatedness evaluations, ConceptNet Numberbatch 16.09
(the complete system described in this paper) is state of the art, performing
better than all other systems evaluated to an extent that exceeds the
confidence interval of the choice of questions. Its high scores on both the
Rare Words dataset and the crowd-sourced MEN-3000 and MTurk-771 datasets,
exceeding the performance of other embeddings with high confidence, shows both
the breadth and the depth of its understanding of words.

\subsection{SAT Analogies}

ConceptNet Numberbatch performed the best among the word-embedding systems at
SAT analogies, getting 56.1\% of the questions correct (58.8\% on the half that
was held out for final testing).  These analogy results outperform analogies
based on other word embeddings, when evaluated in the same framework, as shown
by Figure~\ref{eval-results}.

\begin{table}[t]
\centering
\begin{tabular}{lrrr}
\bf Analogy-solving system   & \bf Accuracy & \bf 95\% conf. \\
\hline
    BagPack (2009)           & .441 & .390 -- .493 \\
    word2vec (2013)          & .486 & .436 -- .537 \\
    SuperSim (2013)          & .548 & .496 -- .599 \\
    LRA (2006)               & .561 & .510 -- .612 \\
    ConceptNet Numberbatch   & .561 & .510 -- .612
\end{tabular}
    \caption{The accuracy of different techniques for solving SAT analogies,
    including ConceptNet Numberbatch 16.09, our hybrid system.}
\label{analogy-eval-table}
\end{table}

The analogy results also tie or slightly outperform the performance of
best-in-class systems on this evaluation\footnote{See
\url{http://www.aclweb.org/aclwiki/index.php?title=SAT_Analogy_Questions} for a
thorough list of results.}. Table~\ref{analogy-eval-table} compares our results
to the other systems introduced in the ``Solving SAT-Style Analogies'' section:
BagPack \cite{herdagdelen2009bagpack}, the previous use of ConceptNet on this
evaluation; LRA \cite{turney2006lra}, the system whose record has stood for a
decade, which spends nine days searching the Web during its evaluation; and
SuperSim \cite{turney2013supersim}, the more recent system that held the record
among self-contained systems. We also include the optimized results we found
for word2vec \cite{mikolov2013word2vec}, which scored best among other
word-embedding systems on this task.

The results of three systems -- SuperSim, LRA, and our ConceptNet Numberbatch
-- are all within each other's 95\% confidence intervals, indicating that the
ranking of the results could easily change with a different selection of
questions. Our score of 56.1\% is also within the confidence interval of the
performance of the average human college applicant on these questions, said to
be 57.0\% \cite{turney2006lra}.

We have shown that knowledge-informed word embeddings are up to the challenge
of real SAT analogies; they perform the same as or slightly better than
non-word-embedding systems on the same evaluation, when other word embeddings
perform worse. In practice, recent word embeddings have instead been evaluated
on simpler, synthetic analogy data sets \cite{mikolov2013word2vec}, and have
not usually been compared to existing non-embedding-based methods of solving
analogies.

We achieve this performance even though the system, like other systems that
form analogies from word embeddings, is only adding and subtracting values that
measure the relatedness of terms; it uses no particular representation of what
the relationships between these terms actually are. There is likely a way to
take ConceptNet's relation labels into account and perform even better at
analogies.

\subsection{Story Cloze Test}

The performance of our system on the Story Cloze Test was acceptable but
unremarkable. ConceptNet Numberbatch chose the correct ending 59.4\% of the
time, which is in fact slightly better than any results reported by
\citeauthor{mostafazadeh2016cloze} \shortcite{mostafazadeh2016cloze}, including
neural nets trained on the task. However, we could also achieve a similar score
by using the same bag-of-vectors approach on other word embeddings. The best
score of 59.9\% was achieved by LexVec, with ConceptNet Numberbatch, GloVe, and
word2vec all within its confidence interval.

This result should perhaps be comforting to those who aim to improve the
computational understanding of stories. A bag-of-vectors approach may be
marginally more successful at choosing the correct ending to a story than other
approaches, but the performance of this approach has likely reached a plateau.
It seems that any sufficiently high-quality word embeddings can choose the
correct ending about 59\% of the time, based on nothing but the assumption that
the end of a story should be similar to the rest of it. Consider this a
baseline: any representation designed to usefully represent the events in
stories should get more than 59\% correct.

\subsection{Conclusion}

We have compared word embeddings that represent only distributional semantics
(word2vec, GloVe, and LexVec), word embeddings that represent only
relational knowledge (ConceptNet PPMI), and the combination of the two
(ConceptNet Numberbatch), and we have shown that the whole is more than the sum
of its parts.

ConceptNet continues to be important in a field that has come to focus on word
embeddings, because word embeddings can benefit from what ConceptNet knows.
ConceptNet can make word embeddings more robust and more correlated with human
judgments, as shown by the state-of-the-art results that ConceptNet Numberbatch
achieves at matching human annotators on multiple evaluations.

Any technique built on word embeddings should
consider including a source of relational knowledge, or starting from a
pre-trained set of word embeddings that has taken relational knowledge into
account. One of the many goals of ConceptNet is to provide this knowledge in a
convenient form that can be applied across many domains and many languages.

\section{Availability of the Code and Data}

The code and documentation of ConceptNet 5.5 can be found on GitHub at
\url{https://github.com/commonsense/conceptnet5}, and the knowledge graph can
be browsed at \url{http://conceptnet.io}.  The full build process, as well as
the evaluation graph, can be reproduced using the instructions included in the
README file for using Snakemake, a build system for data science
\cite{koster2012snakemake}, and optionally using Docker Compose to reproduce
the system environment.  The version of the repository as of the submission of
this paper has been tagged as {\tt aaai2017}.

The ConceptNet Numberbatch word embeddings that resulted from this build
process in September 2016 are the ones evaluated in this paper; they can be
downloaded as pre-built embeddings from
\url{https://github.com/commonsense/conceptnet-numberbatch}, tag {\tt 16.09}.

\section{Acknowledgments}

We would like to thank the tens of thousands of volunteers who provided the
crowd-sourced knowledge that makes ConceptNet possible. This includes
contributors to Open Mind Common Sense and its related projects, as well as
contributors to Wikipedia and Wiktionary, who are improving the state of
knowledge for humans and computers alike.

\bibliographystyle{aaai}
\bibliography{speer-conceptnet}

\end{document}